\documentclass[sigconf]{acmart}
\usepackage{amsmath,amsfonts,physics}
\usepackage{algorithmic}
\usepackage[ruled,linesnumbered,noend]{algorithm2e}      % microtypography
\usepackage{graphicx}
\usepackage{textcomp}
\usepackage{xcolor}
\def\BibTeX{{\rm B\kern-.05em{\sc i\kern-.025em b}\kern-.08em
    T\kern-.1667em\lower.7ex\hbox{E}\kern-.125emX}}
\usepackage[normalem]{ulem}

\usepackage{subfigure}

\newcommand{\scr}[1]{\ensuremath{\mathcal{#1}}}
\AtBeginDocument{%
  \providecommand\BibTeX{{%
    Bib\TeX}}}

\setcopyright{acmcopyright}
\copyrightyear{2022}
\acmYear{2022}
\acmDOI{XXX.XXX}

\begin{document}

\title{Sequence Prediction Under Missing Data : An RNN Approach Without Imputation}

%\author{Paper ID 0301}
%\affiliation{****ReviewVersion****}
%\affiliation{TCS Research}
%\begin{comment}
\author{Soumen~Pachal}
\authornote{Both authors contributed equally to this research.}
\email{s.pachal@tcs.com}
\author{Avinash~Achar}
\authornotemark[1]
\email{achar.avinash@tcs.com}
\affiliation{%
  \institution{TCS Research}
  \streetaddress{IIT Madras Research Park}
  \city{Chennai}
  \state{Tamil Nadu}
  \country{India}
  \postcode{600113}
}
%\end{comment}

\begin{comment}
\author{Lars Th{\o}rv{\"a}ld}
\affiliation{%
  \institution{The Th{\o}rv{\"a}ld Group}
  \streetaddress{1 Th{\o}rv{\"a}ld Circle}
  \city{Hekla}
  \country{Iceland}}
\email{larst@affiliation.org}

\author{Valerie B\'eranger}
\affiliation{%
  \institution{Inria Paris-Rocquencourt}
  \city{Rocquencourt}
  \country{France}
}

\author{Aparna Patel}
\affiliation{%
 \institution{Rajiv Gandhi University}
 \streetaddress{Rono-Hills}
 \city{Doimukh}
 \state{Arunachal Pradesh}
 \country{India}}

\author{Huifen Chan}
\affiliation{%
  \institution{Tsinghua University}
  \streetaddress{30 Shuangqing Rd}
  \city{Haidian Qu}
  \state{Beijing Shi}
  \country{China}}

\author{Charles Palmer}
\affiliation{%
  \institution{Palmer Research Laboratories}
  \streetaddress{8600 Datapoint Drive}
  \city{San Antonio}
  \state{Texas}
  \country{USA}
  \postcode{78229}}
\email{cpalmer@prl.com}

\author{John Smith}
\affiliation{%
  \institution{The Th{\o}rv{\"a}ld Group}
  \streetaddress{1 Th{\o}rv{\"a}ld Circle}
  \city{Hekla}
  \country{Iceland}}
\email{jsmith@affiliation.org}

\author{Julius P. Kumquat}
\affiliation{%
  \institution{The Kumquat Consortium}
  \city{New York}
  \country{USA}}
\email{jpkumquat@consortium.net}
\end{comment}
\renewcommand{\shortauthors}{Soumen et al.}
%\renewcommand{\shortauthors}{0301 et al.}

\begin{abstract}
  Missing data scenarios are very common in ML applications in general  and time-series/sequence applications are no exceptions.
%	Deep learning  in  general  and  Recurrent  neural  networks(RNNs) in particular have seen high success levels in domains ranging from speech recognition, natural language process-ing, computer vision, time series analysis etc. over the last decade. 
This paper pertains to a novel Recurrent Neural Network (RNN) based solution for sequence prediction under missing data.  Our method is 
distinct from all existing approaches. It tries to encode the missingness patterns in the data directly without trying to impute data either before or during model building. Our
encoding is lossless and achieves compression. It can be employed for both sequence classification and forecasting. We focus on forecasting here in a general context of multi-step
prediction in presence of
possible exogenous inputs.  In particular, we propose novel variants of Encoder-Decoder (Seq2Seq) RNNs for this. The encoder here adopts the above mentioned pattern 
encoding, while
	at the decoder which has a different structure, multiple variants are feasible. We demonstrate the utility of our proposed architecture via multiple experiments on both 
	single and  multiple sequence
	(real) data-sets. We  consider both scenarios where (i)data is naturally missing and (ii)data is synthetically masked.
\end{abstract}



\begin{comment}
\begin{CCSXML}
<ccs2012>
 <concept>
  <concept_id>10010520.10010553.10010562</concept_id>
  <concept_desc>Computer systems organization~Embedded systems</concept_desc>
  <concept_significance>500</concept_significance>
 </concept>
 <concept>
  <concept_id>10010520.10010575.10010755</concept_id>
  <concept_desc>Computer systems organization~Redundancy</concept_desc>
  <concept_significance>300</concept_significance>
 </concept>
 <concept>
  <concept_id>10010520.10010553.10010554</concept_id>
  <concept_desc>Computer systems organization~Robotics</concept_desc>
  <concept_significance>100</concept_significance>
 </concept>
 <concept>
  <concept_id>10003033.10003083.10003095</concept_id>
  <concept_desc>Networks~Network reliability</concept_desc>
  <concept_significance>100</concept_significance>
 </concept>
</ccs2012>
\end{CCSXML}

\ccsdesc[500]{Computer systems organization~Embedded systems}
\ccsdesc[300]{Computer systems organization~Redundancy}
\ccsdesc{Computer systems organization~Robotics}
\ccsdesc[100]{Networks~Network reliability}
\end{comment}


\maketitle
\section{Introduction}
\label{sec:introduction}
%\noindent
%{\bf Sequence prediction:} 
Sequence based prediction (classification \cite{Xing10} and forecasting \cite{Lim21}), though an  old research area  continues to be very relevant given recent improvements and constant emergence of 
newer applications.  RNNs are arguably one of the most popular state-of-art sequence predictive modeling approaches. 
Deep RNNs have achieved astounding success over the last decade in domains like NLP, speech and audio processing \cite{Aaron16}, computer vision \cite{Wang16},
time series 
classification, forecasting and so on.  In particular, it has achieved state-of-art performance (and beyond) in  tasks like handwriting recognition
\cite{Graves09}, speech recognition~\cite{Graves13,Graves14}, machine translation~\cite{Cho14,Sutskever14} and   image captioning~\cite{Kiros14,Xu15}, to name
a few. 
%A salient feature in all these applications  is the sequential aspect of the data.

Sequence Prediction under missing data also has a long literature. Missing data could arise due to a variety of reasons like sensor malfunction, maintenance OR even high noise (noise level could be so high that considering data as missing would be more meaningful). Researchers have provided a variety of techniques for data imputation over the years \cite{Johnson13,Debashis10,Koren09,Kira11,Garc10}, which is typically followed by the
predictive modelling step.  RNNs in particular have also been 
explored for sequence modelling under missing data  over the years.
%prediction under missing sequential data over the years. 
This paper addresses sequence prediction under missing data  using RNNs in a novel way. While our proposed ideas can be employed on both classification and forecasting, we focus on
forecasting in this paper. In particular, we consider a general multi-step forecasting scenario with possibly additional exogenous inputs to be handled. 

Encoder-Decoder (ED) OR Seq2Seq architecture used to map variable length sequences to another variable length sequence were first successfully applied for machine translation
\cite{Cho14,Sutskever14} tasks.  From then on, the ED framework has been successfully applied in many other tasks like speech recognition\cite{Liang15},
image captioning etc. Given its variable length Seq2Seq mapping ability, the ED framework has also been utilized for accurate multi-step (time-series) 
forecasting where the target vector length can be variable and independent of the input vector \cite{Wen17}. The missing data RNN architecture proposed here for forecasting is essentially built on the basic ED architecture.  

The forecasting task considered here involves predicting one or more endogenous variables over a multi-step forecast horizon in the presence of possible exogenous inputs which influence the evolution of the endogenous variables. Such forecasting tasks arise in diverse applications. A retailer may be interested in forecasting one or more products sales (modelled as 
endogenous variables) in the presence of exogenous price variation. In electricity markets, forecasting power demand (endogenous) in different geographical regions by factoring temperature influences/fluctuations (exogenous) is 
another example. We consider multi-step forecasting in such scenarios in the presence of missing data.

\smallskip
\smallskip

\subsection{Contributions}
The traditional approach for predictive modelling under missing data has been to impute first and then perform model learning of one's choice.  
Accordingly, there have  been a wide variety 
of data imputation techniques proposed in literature. However, this sequential  approach of impute and model building is very sensitive to the quality of imputation.  
Also, the imputation 
step itself can be computationally intensive. To circumvent the above limitations of this approach, researchers have proposed approaches where imputation and model learning
are carried out jointly.   
Among RNN based missing data works, many approaches follow the jointly impute and learn approach \cite{Bengio95,Che16}.

Our proposed method in this paper adopts a fundamentally different (or distinct) approach compared to the above two broad class of approaches. We try to avoid imputation 
almost completely either before OR during model building.   Our approach performs predictive model building by  directly trying to learn the missingness patterns
observed in the data. This novel general approach can be particularly useful in scenarios where the sequential data is missing in large consecutive chunks. In such scenarios, the existing impute based approaches will tend to suffer significant imputation errors.   

%However, when the sequential data is missing in large consecutive chunks, both these broad class of approaches will tend to fail. Our proposed
%approach is motivated by such considerations and also is distinct from the
%above two broad class of approaches. We try to do model building avoiding imputation to the extent possible. Our approach tries to directly learn the missingness patterns
%observed in the
%data as impute based approaches may not help when data is missing in long chunks. 
%It bases its predictions on the available data and the missingness pattern in the input.  
%Perhaps the only work that comes close to our broad approach of encoding the missingness pais the 
%Our proposed ED
%architecture adopts this broad approach while there are very few works along this line of attack \cite{Lipton16}.

Our contribution involves a novel ED architecture using two encoders, while allowing for intelligent choices/variants in the decoder which  incorporates multi-step (target) learning. 
 Overall  contribution summary is as follows:
\begin{itemize}
	\item We propose a novel encoding scheme of the input window without imputation, which is applicable for both sequence classification and forecasting.   
	For multi-step forecasting in presence of exogenous inputs, we propose a novel ED based architecture which employs the above encoding scheme in the encoder.  
	    \item The missingness pattern in the input window of an example is encoded as it is (without imputation) using two encoders with variable length inputs. 
	    The encoding is lossless with significant compression feasible.
	    %and can possibly reduce the vanishing gradient effect 
	    The output window information can be translated into the decoder in multiple interesting ways. 
%can be substantial depending on the structure of the missingness pattern. 	
	\item  To utilize the proposed scheme for multiple sequence data, where per sequence data is less, we propose a heuristic procedure. 
		%of grouping normalized sequences and build a single (scaled) background model which can be used to predict all sequences.
	 We demonstrate effectiveness of our architecture variants on diverse real data sets.
%(i) real demand data from Australian energy market (ii) real sales data from multiple sources, where the data comes as multiple sequences. 
%To validate the missing data architecture, we synthetically simulated data masking on one of the above  data sets, while additionally tested it on a sales dataset where data 
%was naturally missing.   
%Our experiments illustrate that the proposed methods'
%performance is competitive with state-of-art and outperforms some of the existing methods.  
\end{itemize}
\begin{comment}
(1)by simple imputation of the missing targets and exogenous inputs, (2)additionally encoding the missingness  using a binary encoding (3)unfold the decoder 
(variable length) only till the first missing data point in the
o/p window, (4)joint impute and learn strategy on the exogenous inputs based on  GRU-D, a state-of-art RNN based approach  
\end{comment}

\begin{comment}
The rest of this paper is organized as follows. Section~\ref{sec:Model} describes the DBN model employed in detail and introduces the proposed NoisyOR CPD. Section~\ref{sec:Learning} describes
the learning based on EM idea  and elaborates on how the M-step can be efficiently performed with the proposed NoisyOR CPD.
Section~\ref{sec:Simulation} describes
the validation of the proposed algorithm on synthetic data.
Section~\ref{sec:RelWork} places the contribution of this paper in context of current literature.  
Section~\ref{sec:Conclusions} provides the concluding remarks. 
\end{comment}

\section{Related work}
\label{sec:RelWork}
 Prediction under missing data is a classic problem with an old literature. One broad class of methods to address this is to first impute  followed
by predictive modeling. The simpler approaches towards
 data imputation include smoothing, interpolation, splines\cite{Johnson13} which capture straightforward patterns in the data. The more sophisticated approaches include
 spectral analysis~\cite{Debashis10}, matrix factorization~\cite{Koren09}, kernel methods~\cite{Kira11}, EM algorithm~\cite{Garc10} etc. These sophisticated approaches 
 can  be computationally expensive. Imputation
 methods typically assume restricted scenarios like low missing rates, missing at random and so on which may not hold in practice.
%(please refer to App. \ref{app:DT} for a brief overview)
%For a brief overview of imputation, please refer to App. \ref{app:DT}.
The two step sequential process  of imputation followed by prediction
can suffer from imputation errors while the predictive model so built is blind to any missingness pattern structure.

 RNNs have been explored earlier for time-series tasks (sequence classification in particular) with missing data. The early RNN approaches have predominantly adopted a 
 strategy of jointly imputing and model building. One of the first approaches based on non-gated RNNs seems to be in \cite{Bengio95}. The missing
 values here are initialized to an unconditional expected value and the RNN is unfolded in time for a few extra steps (allowed to relax) to allow the
 missing values to settle to something more reasonable while the other inputs are clamped to their observed values.  The learning criterion is an output 
 error minimization based on all time steps in the unfolded RNN.   In \cite{Tresp97}, to mitigate errors while predicting in a free running OR teacher forcing mode, 
a linear correlated error model is assumed in addition to the NARX model.  The overall learning scheme here involves a mix of Real-time Recurrent Learning for 
the nonlinear map and an EM algorithm for the error model. An RNN   approach  for speech recognition under missing data (on account of noise) was proposed in \cite{Parveen01} which uses an adaption of the architecture in \cite{Bengio95}. In particular, \cite{Bengio95} uses a Jordan network \cite{Jordan88} with feedback or recurrent connections from output to hidden units. In contrast, \cite{Parveen01} avoided feedback from the output but instead  considered recurrent connections from the hidden layer to the input and itself (with a unit-delay) on the lines of an Elman network \cite{Elman90}.

Among the recent RNN approaches, \cite{Lipton16} considers  diagnosis classification under (irregularly recorded) clinical time series. It learns missingness patterns without  any explicit 
 imputation. It encodes the missingness pattern with a simple additional binary vector (indicating presence or absence of data) as input while
training.
More recently, an approach GRU-D \cite{Che16} which incorporates both  missingness pattern (as in \cite{Lipton16}) and the joint impute and learning strategy was proposed.
Here, the missing data points are
assumed to be a convex combination of the last observed value and the unconditional mean. The key intuition is that weight on the last observed value drops (decays) as the time
point 
under consideration moves away from the last observed value (OR time delay). This decay factor is parameterized via a simple perceptron like nonlinear 
function of 
this time-delay whose weights are jointly learnt with the RNN (GRU) weights.
%needed for classification.

Among the recent RNN based forecasting approaches, \cite{CinarNC18} proposes an attention-based encoder-decoder approach of imputing first either using padding (based on left end) OR interpolation (one can use any technique) and then learning by incorporating suitable re-weighting of the impact of these imputed values. This approach doesn't handle exogenous inputs. 
%The standard attention mechanism is  extended by learning some additional position specific weights, dependent on the time difference between prediction instant and position in input window. 
The first missing data variant first imputes using padding from the left-end and considers learning another weight, which models the influence of these imputed values  using an exponential decay based on the distance from the last available data point. The second variant uses interpolation (based on both left and right end values) to impute and then employs  a re-weighting scheme using both the left and right end. It divides a gap into $3$ regions uniformly and learns a weight for each of these regions. 
%Both these variants' re-weighting depends on position 'j' in the input window history.

\subsection{Proposed architecture in perspective}
Our overall approach is to avoid the impute first to the extent possible.  The proposed ED idea employs a unique approach of encoding the missingness pattern in the input
window as it is without any imputation. Instead of using a single encoder with a binary  encoding of the missingness pattern as in \cite{Lipton16}, we employ two encoders
resulting in a compressed and lossless encoding. {\em This feature of our approach is very different  from all existing approaches to the best of our knowledge.}

Our novel  idea of encoding the input window cannot be employed on the output window. This is because  the decoder is unfolded exactly to the extent of forecast
horizon and each time-step of the decoder is used to separately forecast for a  particular future time-step.    However, as we discuss later, one can encode the missing 
exogenous inputs using the binary encoding
of \cite{Lipton16}, which is one of our proposed variants on the decoder. A more powerful variant would be to employ the GRU-D approach on the decoder with exogenous future
variables as inputs.  
\begin{comment}
This can also be strengthened by 
GRU-D \cite{Che16} approach of jointly  impute and learn. We propose to enhance the GRU-D idea of imputing using only the left-end of a gap of inputs and employ it on the
exogenous inputs of the decoder. In the above variants of the proposed approach, the targets in the multi-step output window are imputed. Please note in a forecasting
context, the targets come from the endogenous variable itself. To avoid an impute first strategy of the targets, we propose the jointly impute and learn on the target gaps
as well. 
\end{comment}
 
In summary, our approach incorporates useful aspects of \cite{Lipton16} and \cite{Che16} on the decoder, while it retains its unique distinguishing feature on the encoder side. 

Compared to multi-step forecasting approach of \cite{CinarNC18}, our approach can handle multi-step prediction in the presence of exogenous inputs also. Further, we  avoid
any imputation of any of the input-window missing values unlike \cite{CinarNC18}, by encoding the missingness pattern directly. While \cite{CinarNC18} tries to mitigate the
imputation errors by intelligently
learning to weight the influence of the imputed values towards prediction, it can potentially still  amplify the imputation errors.

\begin{comment}
Moreover, the first variant of \cite{CinarNC18} ignores the right end of any gap of missing values, which can be misleading when the missing points are closer to the right end.  While the second variant even though factors the right-end of a gap, coarsely partitions any gap into 3 regions. This means the influence of an imputed input value is coarsely modelled in the second variant.  
\end{comment}

\section{Proposed Methodology}
\label{sec:Methodology}

Amongst the three standard recurrent structure choices of plain RNN (without gating), LSTM~\cite{Hochreiter97} and GRU~\cite{Chung14}, we choose the GRU
in this paper. Like the LSTM unit, the GRU also has a gating mechanism to mitigate vanishing gradients and have more persistent memory. But the 
lesser gate count in GRU keeps the  number of weight parameters much smaller. GRU unit as the building block for RNNs is currently 
ubiquitous across sequence prediction
applications \cite{Gupta17,Ravanelli18,Che16,Nicole20}.
A single hidden layer plain RNN unit's hidden state can be specified as 
\begin{equation}
	h_t = \sigma(W^{h} h_{t-1} + W^{u} u_t + b)
\end{equation}
where $W^{h}$, $W^{u}$ are weight matrices associated with the state at the previous time-instant $h_{t-1}$ and the current input ($u(t)$) respectively,
$\sigma(.)$ denotes sigmoid function.
GRU based cell computes its hidden state (for one layer as follows)
\begin{eqnarray}
	z_t & = &\sigma(W^z u_t + U^z h_{t-1} + b_z) \label{eq:UpdateGate} \\
	r_t & = & \sigma(W^r u_t + U^r h_{t-1} + b_r) \label{eq:ResetGate} \\
	\tilde{h}_t & = &tanh(U(r_t \odot  h_{t-1})  + W u_t + b) \label{eq:IntermHidState} \\
	h_{t} & = & (1 - z_t) \odot h_{t-1} + z_t\odot \tilde{h}_t
	\label{eq:FinalHidState}
\end{eqnarray}

where $z_t$ is update gate vector and $r_t$ is reset gate vector. If the two gates were absent, we essentially have the plain RNN. $\tilde{h}_t$ is
the new memory (summary of all inputs so far) which is a function of $u_t$ and $h_{t-1}$, the previous hidden state. The reset signal controls 
influence of the previous state on the new memory. The final current hidden state is a convex combination (controlled by $z_t$) of the new memory and memory at the previous
step, $h_{t-1}$. All associated weight matrices $W^z$, $W^r$, $W$, $U^z$, $U^r$, $U$  and vectors $b_z$, $b_r$ and $b$ are trained using back-propagation through time (BPTT).

\subsection{ED Architecture for Missing Data}
\label{sec:Imputation}
\begin{figure*}[!t]
\center
%\includegraphics{../../SIAM/SDMFiles/PlotsSDM/AAAI19FigData/EKFVsPFMAPE.pdf}
%\begin{tabular}{cc}
 %\includegraphics[width=7.0in, height=3.0in]{./Figures/ZeroSalesImputePaper.png}
 \includegraphics[width=7.0in, height=3.6in]{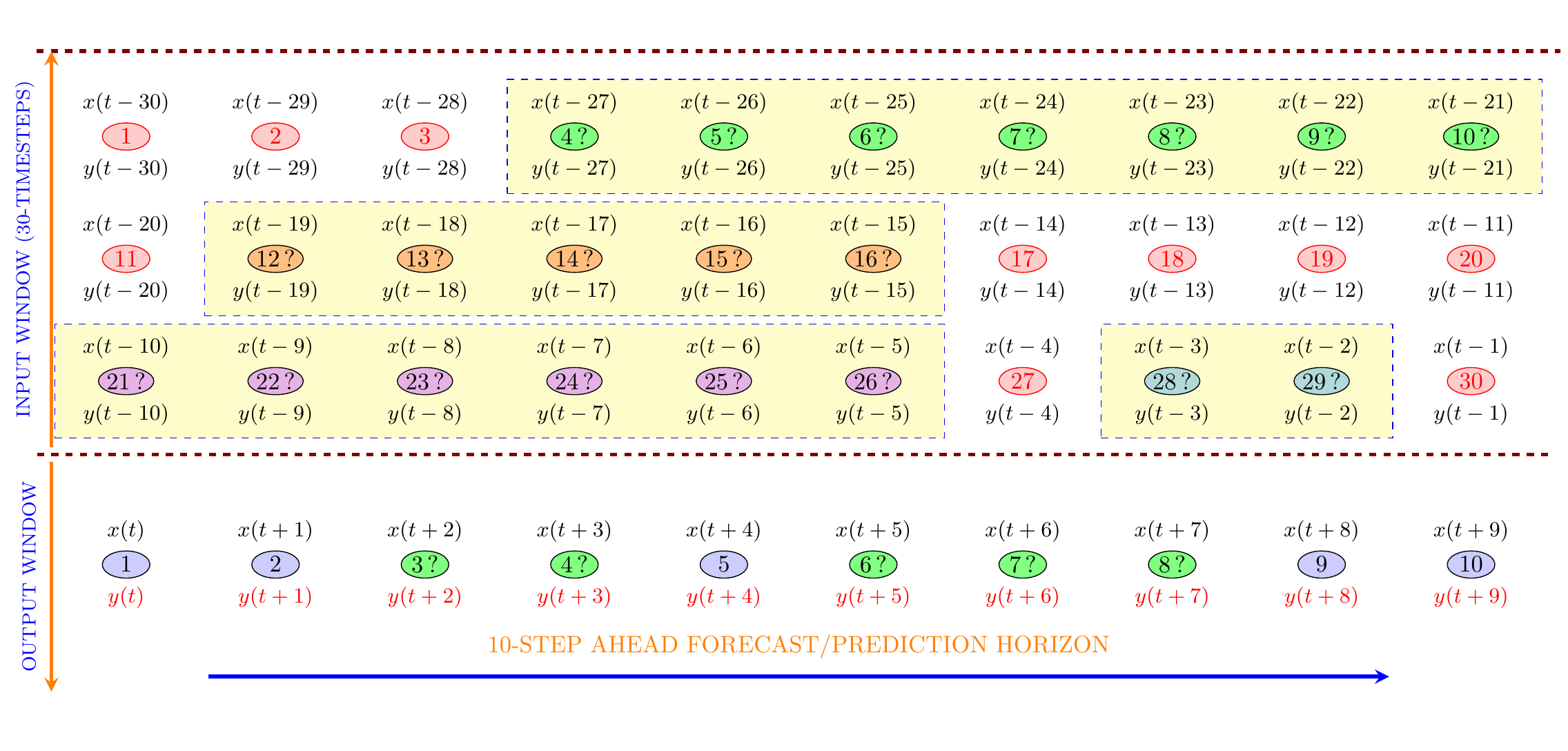}
%\subfigure[RunTime Comparison]{\includegraphics[width=1.6in,height=1.2in]{../SIAM/SDMFiles/PlotsSDM/AAAI19FigData/EKFVsPFRunTime.pdf} \label{fig:RunTimeEKFvsPF}}
%\end{tabular}
	\caption{Illustration of an Example Input-Output Window at time $t-1$ of the sequence data.}
\label{fig:ZSWI}
%\vspace{-0.10in}
\end{figure*}
\begin{figure*}[!t]
\center
%\includegraphics{../../SIAM/SDMFiles/PlotsSDM/AAAI19FigData/EKFVsPFMAPE.pdf}
%\begin{tabular}{cc}
 %\includegraphics[width=7.0in, height=3.0in]{./Figures/ZeroSalesImputePaper.png}
 \includegraphics[width=7.0in, height=2.6in]{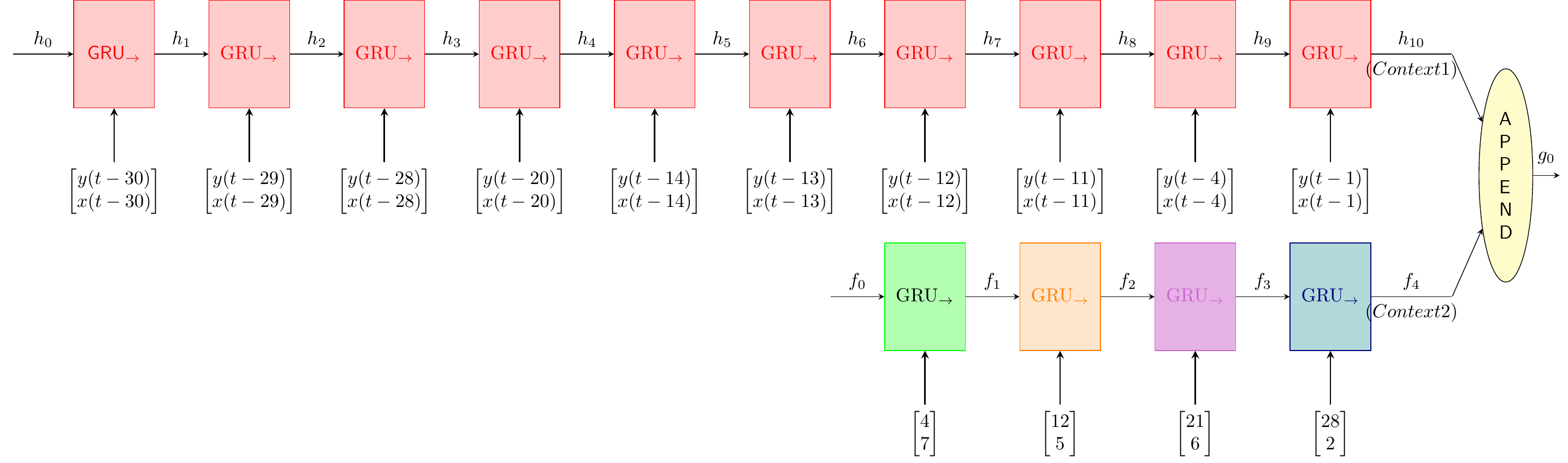}
%\subfigure[RunTime Comparison]{\includegraphics[width=1.6in,height=1.2in]{../SIAM/SDMFiles/PlotsSDM/AAAI19FigData/EKFVsPFRunTime.pdf} \label{fig:RunTimeEKFvsPF}}
%\end{tabular}
	\caption{Illustration of the  Encoder side of the RNN architecture with reference to Example Sequence of Fig.~\ref{fig:ZSWI}}
\label{fig:Encoder}
%\vspace{-0.10in}
\end{figure*}
\begin{figure*}[!h]
\center
%\includegraphics{../../SIAM/SDMFiles/PlotsSDM/AAAI19FigData/EKFVsPFMAPE.pdf}
%\begin{tabular}{cc}
 %\includegraphics[width=7.0in, height=3.0in]{./Figures/ZeroSalesImputePaper.png}
 \includegraphics[width=7.0in, height=1.9in]{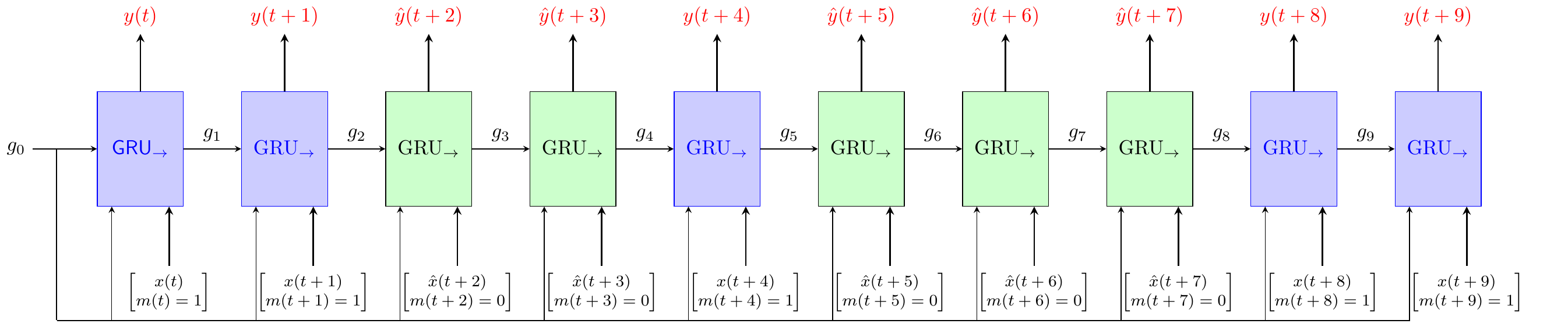}
%\subfigure[RunTime Comparison]{\includegraphics[width=1.6in,height=1.2in]{../SIAM/SDMFiles/PlotsSDM/AAAI19FigData/EKFVsPFRunTime.pdf} \label{fig:RunTimeEKFvsPF}}
%\end{tabular}
	\caption{Illustration of the  Decoder side (Impute with binary encoding) with reference to Example Sequence of Fig.~\ref{fig:ZSWI}}
\label{fig:DecoderFull}
%\vspace{-0.10in}
\end{figure*}

%Missing data scenarios are very common in general in ML applications and time series applications are no exceptions. 
%Missing data could arise due to a variety of reasons like sensor malfunction, maintenance OR even high noise (noise level could be so high that considering data as missing would be more meaningful). 
In this section, we propose a novel 
encoder decoder architecture which can tackle missing data scenarios without imputation (to the extent possible). Most methods for time-series prediction under missing data 
 either (a) impute the data first and then predict (using standard techniques) OR (b) jointly (or simultaneously) learn the best imputation function and final 
 predictive model. In contrast, we propose a general approach which doesn't try to learn the imputation function  as in the above two approaches. It 
 rather tries to learn the final predictive model without imputation with the missing data entries as they are. In other words, it bases its predictions on the available data 
 and the missingness pattern in the input. 
 
Please note while our approach is very general, it can be particularly attractive in certain situations. 
 In some 
time series scenarios, nature of data
missingness could be such that when data is missing, it mostly happens in a medium/large window of consecutive time points. For instance in health care applications, it 
is natural for certain patient health parameters not to be monitored for certain long durations of time depending on the diagnosis.
 In such scenarios imputing the data can be pretty inaccurate especially at time points well into the window. Further if the underlying time series has a high
 total variation, these imputation errors can be more pronounced. Our approach can be particularly useful in such situations. 

\subsubsection{Proposed Encoding}
We consider sequences or time series where the time of occurrence is an integer (or natural number). In particular, we denote a time series of length $N$ as follows.
\begin{equation}
    <\left(x(1),y(1)\right),(x(2),y(2)),\dots (x(t),y(t))\dots (x(N),y(N))>
\end{equation}
where $t\in \mathbb{N}$, $y(t)$ is the endogenous variable (could be vector-valued) which needs to be forecasted and $x(t)$ (could be vector-valued) is the exogenous
variable whose values need to be input into the forecast horizon to forecast the endogenous variable $y(t)$. The assumption of time-series with integer/natural number
ticks is necessary for our proposed method. Another key assumption around the type of missingness is to assume that a $(x(t),y(t))$ pair is either completely missing or
completely present (all the components). 

\noindent
{\bf Key Idea:}{\em The intuition of our RNN scheme is to encode the missingness pattern in the input window of a training example as it is, without imputation. 
This is achieved in a compressed and lossless fashion by using two encoders intelligently. The idea is that this compact input representation can aid the RNN learn better and
faster.} Fig.~\ref{fig:ZSWI} and \ref{fig:Encoder}
illustrate our overall encoding idea with a general example. The idea is to employ two encoders,  one for the available data and the other for the missing data points. The available data points 
in the input time window are fed in the order of their occurrence into the first encoder in spite of the points not being consecutive.   Fig.~\ref{fig:ZSWI}, gives an example input-output window at time tick $t$. The input window dimension is $30$ while the output window is of dimension $10$. The exogenous and endogenous variable/value associated with each time tick of the input window are indicated above and below the (appropriately colored) ellipses. The number in the interior of each of the ellipses indicate the position of the time-tick relative to the start of the window (input or output). Time-ticks where data is missing is additionally quantified with a $'?'$.

Note  
there are $10$ available data points in the input window each marked in red. Observe that all these $10$ points are fed sequentially to the first encoder as illustrated in Fig.~\ref{fig:Encoder}. The second encoder
identifies the blocks of missing data in the input window. Each block can be uniquely identified by two fields: (a)Start time of the block with reference to the
start of the input window (b)width of the block. In Fig.~\ref{fig:Encoder}, there are $4$ such missing data blocks. Each block of consecutive points is marked by a unique color (also grouped separately and shaded in yellow). In particular, the first block's relative start position is $4$,
while its width is $7$. These two bits of information which identify block $1$ are fed as inputs to the first time-step of the encoder $2$. In general, {\em the
$(Start time, window width)$ information of the $i^{th}$ missing data block in the input-window of the data is fed at the $i^{th}$ time-step of the encoder $2$.} As illustrated in Fig.~\ref{fig:Encoder}, the (start, width) of the $4$ missing data blocks in input window of Fig.~\ref{fig:ZSWI} are sequentially fed as inputs at $4$ time-steps 
of encoder $2$. Note the color coding of the $4$ missing data blocks being consistent with the respective GRU block colors to clearly demonstrate our proposed architecture.

This rearranged way of feeding input-window information (without imputation) using two encoders is lossless. 
%As one can see, it is an intelligently compressed version of the input window information. 
Further if data unavailability happens in blocks of large width, our  scheme  achieves significant compression. 
%In particular, the larger the width of these  missing data blocks, higher the compression.   

{\bf \em Computational issues:} Note that based on our encoding idea, the input part of a training example gets transformed into two vectors (fed to the two encoders) of variable dimension. We
exploit this variable length information while training so that the two encoders are sequentially unfolded only to the extent needed. {\em In the presence of data 
missing in blocks, our encoding scheme can potentially work with larger input windows in comparison to what an impute and
predict strategy could due to the vanishing gradient issue.}  This is because the number of unfolded steps in the two encoders via our scheme can be significantly lower than the
input-window size. The example in Fig.~\ref{fig:ZSWI} is a case in point where a standard encoder would have up-to $30$ time steps, while our intelligent encoding leads to at most $10$ steps per encoder. {\em From a training  run-time perspective, for the same input window size, the gradient computation 
can be faster (compared to an impute and predict strategy) via our two-encoder approach given the reduced number of steps in the unfolded structure.}

\subsubsection{Decoder Variants}
\label{sec:DecoderVariants}
%While we chose the encoder inputs without data imputation, 
Our above  idea of intelligently encoding the available and missing data into two separate layers cannot be employed on the decoder. This is because the decoder is performing a (possibly variable length) multi-step forecasting, where at each time-step the decoder output is used to make a prediction. Basically one needs to retain the identity of each step in the forecast horizon in the unfolded decoder. 

{\bf Simple Imputation:}Perhaps the simplest way to handle missing entries in the output window is to impute the missing entries first using a computationally cheap technique. This could be achieved by a simple mean/median imputation for instance. Now the decoder could be unfolded with inputs and targets at each time-step. Fig.~\ref{fig:DecoderFull}
 (with $m(t)$ input ignored) illustrates this for example sequence in Fig.~\ref{fig:ZSWI}. {\em Note the imputed  values $x$ and $y$ are denoted as $\hat{x}$ and $\hat{y}$ respectively.}

\begin{figure}[!h]
\center
%\includegraphics{../../SIAM/SDMFiles/PlotsSDM/AAAI19FigData/EKFVsPFMAPE.pdf}
%\begin{tabular}{cc}
 %\includegraphics[width=7.0in, height=3.0in]{./Figures/ZeroSalesImputePaper.png}
 \includegraphics[width=2.5in, height=1.2in]{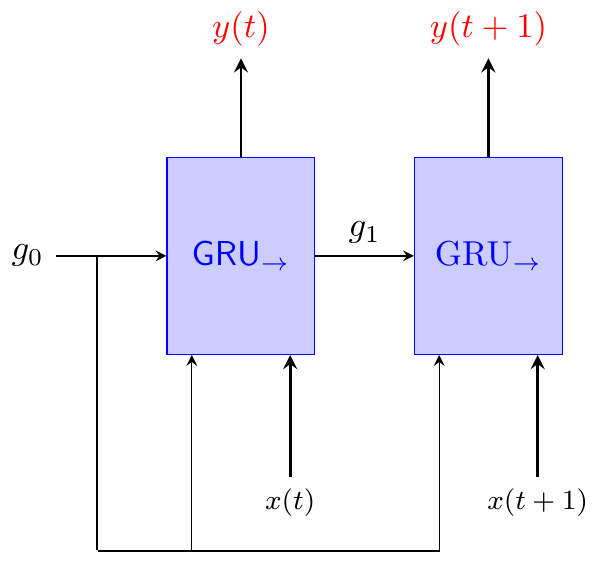}
%\subfigure[RunTime Comparison]{\includegraphics[width=1.6in,height=1.2in]{../SIAM/SDMFiles/PlotsSDM/AAAI19FigData/EKFVsPFRunTime.pdf} \label{fig:RunTimeEKFvsPF}}
%\end{tabular}
	\caption{Variable Length Unfolding in Decoder (based on Fig.~\ref{fig:ZSWI}}
\label{fig:DecoderPartial}
%\vspace{-0.20in}
\end{figure}
{\bf Variable length unfolding:}  A natural way to tackle missing values in output window is to unfold the decoder based on the first set of consecutively available points from  the output window (essentially up to the first missing point). For instance, for example in Fig.~\ref{fig:ZSWI}, we unfold the decoder up to two time-steps only (Fig.~\ref{fig:DecoderPartial}).

{\bf Binary Encoding:} The previous approach while plausible would ignore a training example completely if the target at the first time-step of the output window is missing. But this can lead to severe loss of
examples. A more intelligent approach would be to additionally introduce a binary vector indicating the presence or absence of data at a time-tick as in \cite{Lipton16}. This
means for every input-output window, we obtain an example (we don't miss any example) with decoder unfolded for all steps in the output window on each example. The missing values
could be imputed based on simple imputation as described above.  Fig.~\ref{fig:DecoderFull} indicates the decoder structure with the additional binary input vector ($m(t)$) and imputed missing values for the example in Fig.~\ref{fig:ZSWI}.

\begin{comment}
The decoder is unfolded into as many steps as the prediction
horizon length. If the target at decoder's first time-step is missing, we could ignore such a training example completely. But this can lead to severe loss of
examples. Hence we either choose a zero replacement of the missing values OR to minimally impute based on mean etc. We then use a fully $K$
step time-unfolded 
decoder with appropriate (exogenous) inputs and (target) outputs. 
\end{comment}

{\bf GRU-D based Decoder:}
As explained earlier in related work, GRU-D \cite{Che16} in addition to using the masking binary vector also adopts a novel joint impute-learn strategy.

We define formally  the binary masking variable,  $m_t$, as 
\begin{equation}
 m_t= 
\begin{cases}
    1,& \text{if } x_t \text{ is observed}\\
    0,              & \text{otherwise}
\end{cases}
	\label{eq:BinMask}
\end{equation}

%We first describe our proposed strategy GRU-DE, which is an extension of of GRU-D.  
GRU-D also maintains a time-interval  $\delta_t$,   denoting the distance from its last observation. 
%We maintain two time intervals $\delta_t^L$, $\delta_t^R$ denoting the distance from the closest available data-point on the left and right respectively.
Formally, for integer time-ticks,
\begin{equation}
 \delta_t= 
\begin{cases}
    1 + \delta_{t-1},& \text{if } t>1, m_{t-1}=0\\
    1,              & t>1, m_{t-1}=1\\
    0, &t=1.
\end{cases}
\end{equation}

GRU-D which essentially models a {\em decay} mechanism based on the last observation's distance. The decay factor is modelled for each input variable using a monotonically decreasing function of $\delta_t^L$ ranging between $0$ and $1$.
\begin{equation}
    \gamma_t = exp\{-max(0,W_{\gamma}\delta_t + b_{\gamma})\}
    \label{eq:InputDecay}
\end{equation}
 The modified input that is input to the GRU unit is now 
\begin{equation}
%    \hat{x}_t^d = m_t x_t^d + (1 - m_t)(\gamma_t^d x_t^{dL}  + (1 - \gamma_t^d)\tilde{x}^d)
    \hat{x}(t) = m_t x(t) + (1 - m_t)(\gamma_t x^{L}(t)  + (1 - \gamma_t)\tilde{x})
\label{eq:convex}
\end{equation}
where $x^{L}(t)$ is the last observation to the left of $t$ and $\tilde{x}$ is the empirical mean of $x(t)$.
Replacing $u(t)$ by $\hat{x}(t)$ in eqns.~(\ref{eq:UpdateGate})-(\ref{eq:FinalHidState}) is what a GRU-D recurrent unit is by incorporating input decay alone.
{\em Enhancing the architecture of Fig.~\ref{fig:DecoderFull} using modified exogenous inputs based on eqns.~(\ref{eq:BinMask})-(\ref{eq:convex})  gives us a GRU-D based decoder.}

\subsection{Training on multi-sequence data}
\label{sec:heuristic}
Building sequence-specific models for multi-sequence data, when per-sequence data is less, may be a poor strategy. It can be further compounded if the exogenous variable
additionally shows very little variation per sequence. We propose a heuristic strategy to adapt our above architecture (primarily for single sequences) to such scenarios. The idea
is to build one common background model which can tackle the per-sequence data sparsity. We perform a sequence-specific scaling of each of the input-output windows (for both the 
endogenous and exogenous variables) depending on the sequence from which a particular input-output window was constructed. These normalized examples are used to train a common
background model. During prediction the model output is re-scaled in a sequence-specific fashion.    

\section{Results}
We first describe the data sets used for testing, followed
by error metrics and hyper-parameters for evaluation, baselines  and
performance results in comparison to them. 
\subsection{Data Sets}
Our first data set D1  validates our proposed approach at a single sequence level with about $4$ sequences. Synthetic masking is employed here. The next two data sets are used to vindicate our approach on
multi-sequence data. In particular, synthetic masking is employed in D3, while D2 has missing data points naturally. We provide more details of these data-sets next.

\begin{itemize}
\item {\bf D1:} M5 data-set is a publicly available data-set from
Walmart and contains the unit sales of different products
on a daily basis spanning 5.4 years. This data is distributed across 12 different aggregation levels. We pick sequences from level $12$ where sales and prices are 
available at a product and store level (lowest level with no sales aggregation).
%these levels, aggregation level 8 contains unit sales of all products, aggregated for each store and category. 
Price is used as exogenous input here. While this level contains little more than $30\,000$ sequences, 
we pick $4$ sequences from here with high total variation (referred to as D1). %to test our architecture. 
%and  among the top $10$  sequences (in terms of their total variation). 
We essentially pick the hardest sequences which exhibit sufficient variation. The total 
variation of a $T$ length sequence $x$ is
defined as
\begin{equation}
	TV = \sum_{i=2}^{T} |(x(i+1) - x(i)|
\end{equation}
We test our architecture separately on each of these $4$ sequences by synthetically masking with large width windows.

\item {\bf D2:} This is weekly sales data at an item level from a brick and mortar retail chain (confidential) of a category of items collected over 
$2$ years. There is data missing here naturally on account of no sales and other factors. Per-sequence data is less while price which is the exogenous variable 
has little variation per sequence.
%In this case, the associated price (exogenous variable) doesn't get recorded as there are no transactions for that item in weeks of zero sales. 
Sequences from this category whose fraction of missing points were anywhere less than $60\%$ were chosen to form $D2$. It consists of $1288$ sequences.
\item {\bf D3:} This is  publicly available  from Walmart\footnote{https://www.kaggle.com/c/walmart-recruiting-store-sales-forecasting/data}. The measurements are weekly sales at a department level of multiple departments across 45 Walmart stores. The  price information was chosen as exogenous variable. 
%In addition to sales, there are other related measurements like CPI (consumer price index), mark-down price etc. which we use as exogenous variables for weekly sales prediction. 
The data is collected across $3$ years and it's a multiple time-series data. The whole data set consists of $2628$ sequences. For the purpose of this paper, {\em we ranked sequences based on the total variation of  the sales and considered the top $20\%$ of
the sequences (denoted as D3) for testing}. 
%We essentially pick the hardest sequences which exhibit sufficient variation. The total 
%variation of a $T$ length sequence $x$ is
%defined as
%\begin{equation}
%	TV = \sum_{i=2}^{T} |(x(i+1) - x(i)|
%\end{equation}
	
%It consists of 397 time-series obtained by assimilating all those items where the $\%$ of zero sale weeks was between $20-50\%$.
%\item {\bf D1:} Weekly sales-price at a deparment level of multiple departments ($\approx 600$) from Walmart across $3$ years.
%\item {\bf D2:} Demand (Temperature) data of $3$ regions (independent series) from National Electricity Market, Australia.
%\item {\bf D3:} Weekly sales-price data of items ($\approx 400$) from a  retail chain (confidential) for {\em low moving items} across 
		%$2$ years. Data missing here on account of no sales in a given week.
\end{itemize}
%For more on data set details, please refer to App.~\ref{app:Data}
%We first describe the data sets used for testing, followed by the error metrics and hyperparameters for evaluation, and the performance results of the two architectures in 
%comparison to some strong base lines.
%We test both architectures on D1 using Alg.~\ref{algo:BM} (App. \ref{sec:GreedyRec}). We test seasonal architecture  on $3$ single time-series of D2, while 
%missing data architecture  is tested  on D3 using Alg.~\ref{algo:BM} (App. \ref{sec:GreedyRec})
		{\em Given D2, D3 are multi-sequence data sets with limited per-sequence data, we employ the heuristic background model building idea described in Sec.~\ref{sec:heuristic}.} over a given single-sequence method.
\subsection{Error metrics and Hyper-parameter choices}
We consider the following two  error metrics.
\begin{itemize}
	\item {\bf MAPE} (Mean Absolute Percentage Error)
	\item {\bf MASE} (Mean Absolute Scale Error\cite{Hyndman06})
\end{itemize}

 The APE
is relative error (RE) expressed in percentage. If $\widehat{X}$ is predicted value, while $X$ is the true value, RE = $(\widehat{X} - X)/X$. In
the multi-step setting, APE is computed for each step and is averaged over all steps to obtain the MAPE for one window of the prediction horizon.
APE while has the advantage of being  a scale independent metric, can assume abnormally high values and can be misleading when the true value is very low. An
alternative complementary error metric which is scale-free could be MASE. 

The MASE is computed with reference to a baseline metric. The choice of baseline is
typically the {\em copy previous} predictor, which just replicates the previous observed value as the prediction for the next step.  
For a given window of one prediction horizon of $K$ steps ahead, let us denote the $i^{th}$ step error by $|\widehat{X}_i - X_i|$. The $i^{th}$ scaled error is
defined as 
\begin{equation}
	e_s^i =  \frac{|\widehat{X}_i - X_i|} { \frac{1}{n-K}\sum_{j=K+1}^{n}|X_j - X_{j-K}|}
\end{equation}
where $n$ is no. of data points in the training set. The normalizing factor is the average $i^{th}$ step-ahead error of the copy-previous baseline 
on the training set. Hence the MASE on a multi-step prediction window $w$ of size $K$ will be 
\begin{equation}
	MASE(w,K) = \frac{1}{K}\sum_{j=1}^{K} e_s^j
\end{equation}

Table~\ref{tab:HP} describes the broad choice of hyper-parameters during training in our experiments.
\begin{table}[!htbp]
%\begin{center}
\caption{Model parameters during training.}
\label{tab:HP}
\centering
\begin{tabular}{|c |c| }
 \hline
	{\bf Parameters} & {\bf Description} \\ \hline
 Batch size & 64/256   \\ %\hline
 Number of Hidden layers & 1/2   \\ %\hline
 Hidden vector dimensionality  & 7/9    \\ %\hline
 Optimizer & RMSProp / Adam   \\ \hline
\end{tabular}
%\end{center}
\end{table}
%Refer to App. \ref{sec:HP} for hyperparameter details during training.

%%%%%%%%%%%%%%%%%%%%%%%%%%%%%%%%%%%%%%%%%%%%%%%%%%%%

\subsection{Baselines and Proposed Approaches}
\label{sec:WI}
We benchmark two of the discussed decoder variants (Sec.\ref{sec:DecoderVariants}): (a)Simple (Mean) Imputation denoted as DEMI (b)GRU-D based decoder imputation denoted as DEGD. 
DE refers to double encoder or the two RNN layers used to encode input window (Fig.~\ref{fig:Encoder}).
Recall from Sec.~\ref{sec:DecoderVariants} that `variable length unfolding' variant  can suffer from severe loss of examples while  the `Binary encoding' variant is a special case of GRU-D based variant. Hence, we benchmark DEMI and DEGD only. 
The baselines we benchmark our method against are as follows:
\begin{enumerate}
	\item BEDXM - post mean imputation (in all missing points) run a Basic Encoder-Decoder (with one encoder capturing immediate lags) with exogenous inputs as in \cite{Wen17}. 
%	\item  MedI - post median imputation  run a BEDX
	\item BEDXL - Consider any band of missing points.  The center of the band is imputed
		with the mean while from both the left and right end the missing points are now linearly interpolated. Next use \cite{Wen17} as above. It uses a simple imputation method but utilizes both the right and left end information of any gap unlike \cite{Che16}. 
		
	%\item BM - binary encoding of missing pattern \cite{Lipton16}.
	\item GRU-D\cite{Che16} - at both the encoder and decoder, we use the GRU-D unit to achieve multi-step forecasting (something unexplored in the original \cite{Che16} paper).
	\item WIAED\cite{CinarNC18} - which learns the position-based weighted influence of imputed values followed by attention layer on the encoder-decoder framework.   
	%\item MWAI - impute using a moving window average  
\end{enumerate}
\begin{comment}
is imputed using a linear interpolation strategy.
Recall from Sec.~\ref{sec:RelWork} that  \cite{Che16} is doing something simpler
where any missing point is filled as a convex weighted combination of the left end point and the mean with the weight (on left end point) decaying with distance from the left end
point. It totally ignores the right side, while the weight decay learnt from data. Here we assume a linear decay from either side and in this sense our baseline is more 
general than \cite{Che16}.
\end{comment}

\subsubsection{\bf Assessing significance of mean error differences
statistically} We have conducted a Welch t-test (unequal
variance) based significance assessment (across all relevant
experiments) under both the mean metric (MASE, MAPE)
differences (Proposed vs Baseline) with a significance level of
0.05 for null hypothesis rejection. The best performing method's error is 
highlighted in bold if its MASE/MAPE improvement over every other method is statistically significant. We allow for highlighting the second/third best errors in situations when the mean error differences between the best and second best errors are statistically insignificant. 

\subsubsection{\bf Synthetic Masking} The masking process adopted is as follows. We scan the time-series sequentially from the start. At each step, we essentially decide
if a masking has to be carried out starting from the current step.
	 Towards this, we toss a coin with a variable heads probability $q$. If tails, we decide against masking from the current step and move on. On heads, we decide 
to mask from current step.  After seeing a head
	every time,  a decision of how many consecutive data points to mask needs to be taken.
Towards this, we consider  $\scr{T}_{w}$, which is set of window lengths from which we uniformly sample a specific length. Let $T_w$ denote the window
length sampled from $\scr{T}_{w}$.  For the next masking window starting position, we choose $q=1/T_{w}$. This strategy assures that about $50\%$ of the data
points are missing because for every ${T}_{w}$ points removed, we retain the next ${T}_{w}$ points (in expectation).  
Initial $q$ is set to $0.05$.

Synthetic masking has the following advantage. Since the underlying true data before masking is available, 
one can 
assess the efficacy of training on a separate test set for every step of the prediction horizon.
\subsection{Results on D1 (after synthetic masking)}
For D1, the forecast horizon was set to
be $28$ days (K = 28). The choice of $K=28$ would mean about $4$ weeks ahead (which is a meaningful and not an overly long forecast horizon). This was also the forecast horizon of the 
$M5$ challenge. A test size of $127$ days (time-points) out of $1941$ time points was
set aside for each sequence in D1. This means we tested for
$100$ output windows of width $28$  per sequence.

To illustrate on D1 (data is not missing), we artificially simulate  masking using long masking windows.    
Tab.~\ref{tab:MaxAvgMinD2} provides errors of all relevant methods.  Results indicate superior performance of both DEMI and DEGD with best case MAPE improvements per baseline 
of $14\%$, $8\%$, $4\%$ and $28\%$ with respect to BEDXM, BEDXL, GRU-D and WIAED respectively. Similarly, the best case MASE improvements per baseline were $0.17$, $0.16$,
$0.05$  and  $0.39$ respectively. Also note all improvements of the proposed approaches are statistically significant.
 
\begin{table}[!htbp]
\footnotesize
%\small
%\begin{center}
	\caption{D1 results:  $\scr{T}_w = \{30,31,32,\dots,39,40\}$, Input-Window width=$200$.}
\label{tab:MaxAvgMinD2}
\centering
\begin{tabular}{|c ||c|c|c|c|c|c|c|}
 \hline
Seq ID & DEMI & DEGD    & BEDXM	& BEDXL  &  GRU-D&    WIAED\\  \hline
$FOODS\_3\_714$ &{\bf(0.47,38)} &{\bf(0.46,38)}    &(0.56,46) &(0.56,46) &(0.49,41) &(0.48,39) \\ 
$FOODS\_3\_252$ &{\bf (0.51,23)} &{\bf (0.51,23)}    &(0.54,26) &(0.57,25) &(0.55,25)&(0.74,33) \\ 
$FOODS\_3\_080$ &{\bf (0.16,7)} &{\bf (0.15,7)}   &(0.32,21) &(0.31,15) &(0.20,11) &(0.54,35) \\  
$FOOD\_3\_090$ &{\bf (0.29,19)} &{\bf (0.29,19)}    &(0.41,26) &(0.45,27) &(0.32,21)&(0.50,41) \\ \hline
%$FOOD\_3\_090$ &{\bf (0.29,19)} &{\bf (0.29,19)}    &(0.41,26) &(0.45,27) &(0.32,21)&(0.50,41) \\ \hline
\end{tabular}
%\end{center}
\end{table}

\subsection{Results on D2}
\begin{figure}[!htbp]
\center
	 \includegraphics[width=3.4in,height=1.6in]{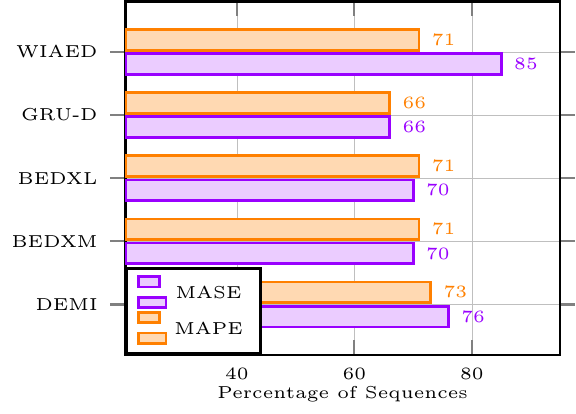}
%\begin{tabular}{cc}
%\subfigure[12 week data]{\includegraphics[width=2.50in,height=1.5in]{../../SIAM/SDMFiles/PlotsSDM/AAAI19FigData/MultiStepMAENew3.pdf} \label{fig:MultStepNew}} 
%\subfigure[5 week data]{\includegraphics[width=2.50in,height=1.5in]{../../SIAM/SDMFiles/PlotsSDM/AAAI19FigData/MultiStepMAEOld3.pdf} \label{fig:MultStepOld}}
%\end{tabular}
	 \caption{Percentage of sequences (out of $940$ on which at least one method's MASE $<1$) where DEGD does better.}
	 \label{fig:FracWIAB}
%\vspace{-0.10in}
\end{figure}
\begin{table}[!htbp]
%\vspace{-0.15in}
%\begin{center}
\caption{Max, Avg and Min of MASE  across all relevant $940$ sequences (on which at least one method's MASE $<1$)}
\label{tab:MaxAvgMinWIAB}
\small
\centering
\begin{tabular}{|c ||c|c|c| }
 \hline
	& \multicolumn{3}{|c|}{MASE based} \\ \cline{2-4}
	Method  & Max	& Avg  &  Min   \\  \hline
	DEMI &21.44 &1.79   &0.004  \\ %\hline
	DEGD  &19.61 &{\bf 0.43}   &0.001  \\ %\hline
BEDXM &17.26 &1.34    &0.001  \\ %\hline
% MedI&  &   & \\ %\hline
 BEDXL &16.98  &1.37    &0.002 \\ %\hline
	GRU-D &16.87 &0.52   &0.0002  \\ %\hline
% BM&8.39  &0.36   &0.001   \\ %\hline
 WIAED &24.11 &0.93   &0.005  \\ \hline
\end{tabular}
%\end{center}
%\vspace{-0.10in}
\end{table}
For D2, input window size is chosen to be $20$, while the prediction horizon chosen was of length $K=12$ (decoder length OR output window size), which is a $3$-month ahead forecast horizon. This choice of input and output window length (per example) results in a reasonably  long forecast horizon while also generating
sufficient number of input-output examples per sequence given that we have about $104$ weeks of data points per sequence. We tested for
$6$ output windows  of width $12$ on a $17$ week (separately kept) test set per sequence.

{\em For about $348$ items,
MASE $>1$, for all baselines and the proposed methods. which meant these items could be tackled better by the simple copy previous baseline.} We hence kept these items aside 
and concentrate on the remaining $940$, on each of which at least one of the baselines OR proposed approaches  have MASE $<1$. 
Fig.~\ref{fig:FracWIAB}  gives a detailed breakup of the percentage of these $940$ sequences on which DEGD did better compared to the $3$ baselines. It demonstrates that DEGD does
better on at least $66\%$ of sequences and up to $85\%$, compared to all baselines.
\begin{comment}
\begin{table}[!htbp]
%\begin{center}
	\caption{Percentage of sequences (out of $274$ on which at least one method's MASE $<1$)}
\label{tab:FracWIAB}
\centering
\begin{tabular}{|c ||c|c| }
 \hline
 Baseline &MASE based & MAPE based   \\ \hline
 MnI &67 & 69   \\ %\hline
 MedI & 61 & 64   \\ %\hline
 LinI & 72 & 75    \\ %\hline
 BM & 71 & 73    \\ \hline
\end{tabular}
%\end{center}
\end{table}
\end{comment}
 
Tab.~\ref{tab:MaxAvgMinWIAB} gives the average, max and min across sequences (of MASE) for all methods.  It demonstrates that on an average DEGD does better than all baselines.
%based on both these complementary metrics. 
{\em MASE improvements are up to $0.94$.}
\begin{comment}
\begin{table}[!htbp]
%\begin{center}
\caption{Max, Avg and Min of MASE and MAPE across all sequences whose MASE $<1$}
\label{tab:MaxAvgMinWIAB}
\centering
\begin{tabular}{|c ||c|c|c|c|c|c| }
 \hline
	& \multicolumn{3}{|c|}{MASE based} & \multicolumn{3}{|c|}{MAPE based}\\ \cline{2-7}
	Method  & Max	& Avg  &  Min & Max	& Avg  &  Min  \\  \hline
 WIEX &1.00 & 0.41  &0.01 & $>$1000 &107 & 3 \\ %\hline
MnI &3.31 & 0.53   &0.02 & $>$1000  &140 & 5 \\ %\hline
 MedI& 1.56 & 0.46  &0.01 & $>$1000&123 & 5 \\ %\hline
 LinI& 2.70 & 0.58   &0.02 & $>$1000&144 & 3 \\ %\hline
 BM& 2.29 & 0.56  &0.02 & $>$1000&150 & 5  \\ \hline
\end{tabular}
%\end{center}
\end{table}
\end{comment}
Tab.~\ref{tab:CondAvgMASEWIAB} looks at the (conditional) average MASE under two conditions with respect to each baseline: (i)average over those sequences on 
which DEGD fares better, (ii)average over those sequences on which the baseline does better.  At this level, {\em we observe MASE improvements of at least $0.22$ while up to
$1.58$.}
%{\em For results on $D1$ post synthetic masking, please refer to App. \ref{sec:MissingD1}.}

\begin{comment}
\begin{table}[!htbp]
%\vspace{-0.10in}
\caption{Average MASE on $274$ relevant sequences when (i)WIEX fares better (ii)Baseline fares better.}
\label{tab:CondAvgMASEWIAB}
\footnotesize
\centering
\begin{tabular}{|c ||c|c|c|c|c|c| }
 \hline
	& \multicolumn{3}{|c|}{WIEX better} & \multicolumn{3}{|c|}{Baseline better}\\ \cline{2-7}
	Method  & WIEX & BLine &  Diff & WIEX & BLine &  Diff \\  \hline
 %SEDX &25 & 15  &25 & 15&25 & 15 \\ \hline
	MnI  &0.41 &0.62   &{\bf0.21} &0.55 &0.48 &0.07  \\ %\hline
	MedI&0.43  & 0.55  &{\bf0.12} &0.51  &0.41 &0.10  \\ %\hline
	LinI&0.40  & 0.69   & {\bf0.29}&0.59 &0.44 &0.15  \\ %\hline
	BM&0.41  &0.67   &{\bf0.26} &0.59 &0.48 &0.11   \\ \hline
\end{tabular}
%\end{center}
%\vspace{-0.10in}
\end{table}
\end{comment}

\begin{table}[!htbp]
%\vspace{-0.10in}
\caption{Average MASE on $940$ relevant sequences when (i)DEGD fares better (ii)Baseline fares better.}
\label{tab:CondAvgMASEWIAB}
\footnotesize
\centering
\begin{tabular}{|c ||c|c|c|c|c|c| }
 \hline
	& \multicolumn{3}{|c|}{DEGD better} & \multicolumn{3}{|c|}{Baseline better}\\ \cline{2-7}
	Method  & DEGD & BLine &  Diff & DEGD & BLine &  Diff \\  \hline
 %SEDX &25 & 15  &25 & 15&25 & 15 \\ \hline
 DEMI  &0.26 &2.18   &{\bf 1.92} &0.97 &0.59 &0.38  \\ %\hline
	BEDXM  &0.21 &1.75   &{\bf 1.54} &0.94 &0.40 & 0.54 \\ %\hline
%	MedI&0.43  & 0.55  &{\bf0.12} &0.51  &0.41 &0.10  \\ %\hline
	BEDL&0.21  &1.79    &{\bf 1.58} &0.93 &0.41 &0.52  \\ %\hline
 GRU-D  &0.24 &0.46   &{\bf 0.22} &0.80 &0.64 &0.16  \\ %\hline
	%BM&0.22  &0.30   &0.08 &0.99 &0.53 &{\bf 0.46}   \\ %\hline
	WIAED &0.34 &0.98   &{\bf 0.64} &0.95 &0.69 &0.26  \\ \hline
\end{tabular}
%\end{center}
%\vspace{-0.10in}
\end{table}

\begin{comment}
\begin{table}[!htbp]
%\begin{center}
	\caption{Average MAPE on sequences when (i)WIEX fares better (ii)Baseline fares better.}
\label{tab:CondAvgMAPEWIAB}
\centering
\begin{tabular}{|c ||c|c|c|c|c|c| }
 \hline
	& \multicolumn{3}{|c|}{WIEX better} & \multicolumn{3}{|c|}{Baseline better}\\ \cline{2-7}
	Method  & WIEX & BLine &  Diff & SEDX & BLine &  Diff \\  \hline
 %SEDX &25 & 15  &25 & 15&25 & 15 \\ \hline
MnI   &1.06  & 1.59  &0.53 & 44 & 19 & 25 \\ \hline
MedI  & 20 & 79  &60 & 46 &20 & 26 \\ %\hline
LinI  & 18 & 82   &64  & 42&20 & 22 \\ \hline
BM & 23 & 33  &10 & 36&23 & 13  \\ \hline
\end{tabular}
%\end{center}
\end{table}
\end{comment}

\subsection{Results on D3 after synthetic masking}
\label{sec:MissingD1}
As explained earlier, we perform a synthetic masking first. Note the per-sequence length here is about $150$ points.  Compared to D1, the input window length  
has to chosen to be comparatively lower (chosen to be $20$ here as in D2) to obtain reasonable number of input-output examples from each sequence. This means the  masking window length
 has to be proportionately less (compared to that of D1). 

A test size of 15 weeks (time-points) was 
set aside for each sequence (out of about $150$ weeks) in D3. We choose K = 10 time-steps in the decoder (output window length) for training  which means forecast time-horizon would be about $2.5$ months, a reasonably long
forecast window which also allows for sufficiently many input output examples from each sequence.
We tested for
$6$ output windows (as in D2) of width $10$ on the $15$ week test set per sequence.

\begin{figure}[!htbp]
\center
	 \includegraphics[width=3.4in,height=1.6in]{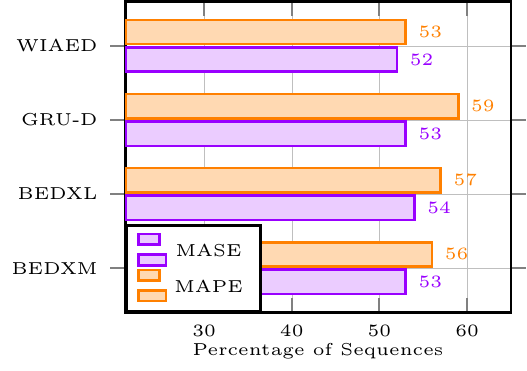}
%\begin{tabular}{cc}
%\subfigure[12 week data]{\includegraphics[width=2.50in,height=1.5in]{../../SIAM/SDMFiles/PlotsSDM/AAAI19FigData/MultiStepMAENew3.pdf} \label{fig:MultStepNew}} 
%\subfigure[5 week data]{\includegraphics[width=2.50in,height=1.5in]{../../SIAM/SDMFiles/PlotsSDM/AAAI19FigData/MultiStepMAEOld3.pdf} \label{fig:MultStepOld}}
%\end{tabular}
	 \caption{Percentage of sequences where DEMI does better. $q=0.05$, $\scr{T}_w = \{6,7,8,9,10\}$.}
	 \label{fig:FracWI}
%\vspace{-0.10in}
\end{figure}

\begin{figure}[!htbp]
\center
	 \includegraphics[width=3.4in,height=1.6in]{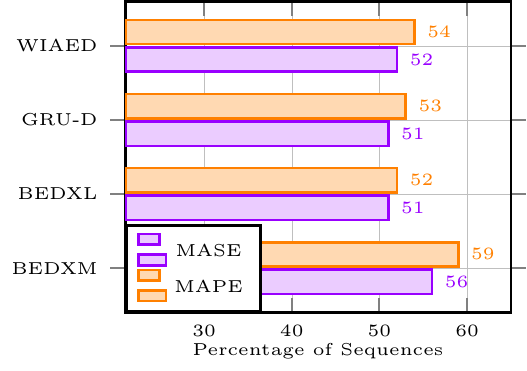}
%\begin{tabular}{cc}
%\subfigure[12 week data]{\includegraphics[width=2.50in,height=1.5in]{../../SIAM/SDMFiles/PlotsSDM/AAAI19FigData/MultiStepMAENew3.pdf} \label{fig:MultStepNew}} 
%\subfigure[5 week data]{\includegraphics[width=2.50in,height=1.5in]{../../SIAM/SDMFiles/PlotsSDM/AAAI19FigData/MultiStepMAEOld3.pdf} \label{fig:MultStepOld}}
%\end{tabular}
	 \caption{Percentage of sequences where DEGD does better. $q=0.05$, $\scr{T}_w = \{6,7,8,9,10\}$.}
	 \label{fig:DEGDComp}
\vspace{-0.10in}
\end{figure}

Fig.~\ref{fig:FracWI}  gives a detailed breakup of percentage of sequences on which DEMI did better compared to the $4$ baselines. {\em It demonstrates that DEMI does better on at least
$52\%$ of the sequences and up to $59\%$ compared to all considered baselines. } A similar plot  is obtained when DEGD is compared with the $4$ baselines as shown in 
 Fig.~\ref{fig:DEGDComp}. %of App.~\ref{sec:ExtraD3}. 
 {\em This clearly indicates similar performance of DEGD and DEMI on D3.}
	\begin{comment}
\begin{table}[!htbp]
%\begin{center}
	\caption{Percentage of sequences where WIEX does better. $q=0.05$, $\scr{T}_w = \{6,7,8,9,10\}$.}
\label{tab:FracWI}
\centering
\begin{tabular}{|c ||c|c| }
 \hline
 Baseline &MASE based & MAPE based   \\ \hline
 MnI &75 & 65   \\ %\hline
 MedI & 77 & 67   \\ %\hline
 LinI & 69 & 59    \\ %\hline
 BM & 64 & 59    \\ \hline
\end{tabular}
%\end{center}
\end{table}
	\end{comment}

Tab.~\ref{tab:MaxAvgMinWI} gives the average, max and min across sequences (of MASE and MAPE) for all methods.  It demonstrates that on an average DEMI does better than all
baselines 
based on both these complementary metrics. MASE improvements are up to $0.10$ while the MAPE improvements are up to $5\%$.
\begin{comment}
\begin{table}[!htbp]
%\begin{center}
\caption{Max, Avg and Min of MASE and MAPE across all sequences}
\label{tab:MaxAvgMinWI}
\centering
\begin{tabular}{|c ||c|c|c|c|c|c| }
 \hline
	& \multicolumn{3}{|c|}{MASE based} & \multicolumn{3}{|c|}{MAPE based in $\%$}\\ \cline{2-7}
	Method  & Max	& Avg  &  Min & Max	& Avg  &  Min  \\  \hline
	WIEX &2.84 & {\bf 0.81}  &0.17 & 379& {\bf 28} & 2 \\ %\hline
 MnI &9.68 & 1.66  &0.14 & 498 &57 & 3 \\ %\hline
 MedI& 8.10 & 1.67  &0.26 & 453&60 & 2 \\ %\hline
 LinI& 8.51 & 1.67   &0.16 & 497&56 & 3 \\ %\hline
 BM& 3.38 & 0.92  &0.16 & 310&29 & 2  \\ \hline
\end{tabular}
%\end{center}
\end{table}
\end{comment}
Tab.~\ref{tab:CondAvgMASEWI} looks at the (conditional) average MASE under two conditions with respect to each baseline: (i)average over those sequences on 
which DEMI fares better, (ii)average over those sequences on which the baseline does better.  At this level, MASE improvements of at least $0.28$ while up to $0.60$ are observed.  
Tab.~\ref{tab:CondAvgMAPEWI} 
%in App.~\ref{sec:ExtraD3} 
considers a similar  (conditional) average MAPE. At this level of MAPE, there are improvements of at least $11\%$ to up to $18\%$. 
 {\em Overall our results on D3 indicate that our approach is viable even when the masking window length is low.}

\begin{comment}
\begin{table}[!htbp]
\caption{Average MASE when (i)WIEX fares better (ii)Baseline fares better.}
\label{tab:CondAvgMASEWI}
\footnotesize
\centering
\begin{tabular}{|c ||c|c|c|c|c|c| }
 \hline
	& \multicolumn{3}{|c|}{WIEX better} & \multicolumn{3}{|c|}{Baseline better}\\ \cline{2-7}
	Method  & WIEX & BLine &  Diff & WIEX & BLine &  Diff \\  \hline
 %SEDX &25 & 15  &25 & 15&25 & 15 \\ \hline
	MnI  &0.73 & 1.96  &{\bf 1.23} & 1.06&0.78 & 0.28 \\ %\hline
	MedI& 0.72 & 1.94  &{\bf 1.22} & 1.10&0.78 & 0.32 \\ %\hline
	LinI& 0.74 & 2.13   &{\bf 1.39} & 0.96&0.66 & 0.30 \\ %\hline
	BM& 0.73 & 1.03  &{\bf 0.30} & 0.95&0.73 & 0.22  \\ \hline
\end{tabular}
%\vspace{-0.10in}
%\end{center}
\end{table}
\end{comment}

\begin{comment}
\begin{table}[!htbp]
\caption{Average MAPE when (i)WIEX fares better (ii)Baseline fares better.}
\label{tab:CondAvgMAPEWI}
\footnotesize
\centering
\begin{tabular}{|c ||c|c|c|c|c|c| }
 \hline
	& \multicolumn{3}{|c|}{WIEX better} & \multicolumn{3}{|c|}{Baseline better}\\ \cline{2-7}
	Method  & WIEX & BLine &  Diff & WIEX & BLine &  Diff \\  \hline
 %SEDX &25 & 15  &25 & 15&25 & 15 \\ \hline
	MnI   &20 & 77  &{\bf 57} & 44 & 19 & 25 \\ %\hline
	MedI  & 20 & 79  &{\bf 59} & 46 &20 & 26 \\ %\hline
	LinI  & 18 & 82   &{\bf 64}  & 42&20 & 22 \\% \hline
	BM & 23 & 33  &{\bf 10} & 36&23 & 13  \\ \hline
\end{tabular}
%\end{center}
\end{table}
\end{comment}

\begin{table}[!htbp]
%\begin{center}
	\footnotesize
\caption{Max, Avg, Min of MASE/MAPE across all sequences}
\label{tab:MaxAvgMinWI}
\centering
\begin{tabular}{|c ||c|c|c|c|c|c| }
 \hline
	& \multicolumn{3}{|c|}{MASE based} & \multicolumn{3}{|c|}{MAPE based in $\%$}\\ \cline{2-7}
	& Max	& Avg  &  Min & Max	& Avg  &  Min  \\  \hline
	DEMI &4.03 &{\bf 0.93}   &0.29 &349 &{\bf 31} &2 \\ %\hline
 DEGD  &3.77 &0.94   &0.08 &397 &32 &0.4 \\ %\hline
 BEDXM &3.62 &0.99   &0.08 &347  &36 &0.9  \\ %\hline
 %MedI&  &   & & & &  \\ %\hline
 BEDXL &3.43  &0.96    &0.03 &382 &35 &0.4  \\ %\hline
 GRU-D &3.53 &0.96   &0.08 &318 &33 &2 \\ %\hline
 %BM&3.40  &0.94   &0.07 &385 &35 &0.7    \\ %\hline
 WIAED &4.58 &1.03   &0.09 &250 &33 &0.7 \\ \hline
\end{tabular}
%\end{center}
\end{table}

\begin{table}[!htbp]
\caption{Average MASE on sequences where (i)DEMI fares better (ii)Baseline fares better.}
\label{tab:CondAvgMASEWI}
\footnotesize
\centering
\begin{tabular}{|c ||c|c|c|c|c|c| }
 \hline
	& \multicolumn{3}{|c|}{DEMI better} & \multicolumn{3}{|c|}{Baseline better}\\ \cline{2-7}
	Method  & DEMI & BLine &  Diff & DEMI & BLine &  Diff \\  \hline
 %SEDX &25 & 15  &25 & 15&25 & 15 \\ \hline
 DEGD  &0.84 &1.12   &0.28 &1.04 &0.75 &{\bf 0.29}  \\ %\hline
	BEDXM &0.84 &1.24   &{\bf 0.40} &1.03 &0.71 &0.32  \\ %\hline
%	MedI&  &   & & & &  \\ %\hline
	BEDXL &0.84  &1.18    &{\bf 0.34} &1.04 &0.71 & 0.33 \\ %\hline
 GRU-D  &0.84 &1.14   &{\bf0.30} &1.03 &0.76 &0.27  \\ %\hline
	%BM&0.83  &1.16   &{\bf 0.33} &1.03 &0.71 &0.32   \\ %\hline
	WIAED &0.86 &1.46   &{\bf 0.60} &1.00 &0.58 &0.42  \\ \hline
\end{tabular}
%\vspace{-0.10in}
%\end{center}
\end{table}

\begin{table}[!htbp]
\caption{Average MAPE on sequences where (i)DEMI fares better (ii)Baseline fares better.}
\label{tab:CondAvgMAPEWI}
\footnotesize
\centering
\begin{tabular}{|c ||c|c|c|c|c|c| }
 \hline
	& \multicolumn{3}{|c|}{DEMI better} & \multicolumn{3}{|c|}{Baseline better}\\ \cline{2-7}
	Method  & DEMI & BLine &  Diff & DEMI & BLine &  Diff \\  \hline
 %SEDX &25 & 15  &25 & 15&25 & 15 \\ \hline
DEGD   &28 &39   &11 &36  &23  &{\bf 13}  \\ %\hline
	BEDXM &29 &47   &{\bf 18} &34  &21  &13  \\ %\hline
%	MedI  &  &   & &  & &  \\ %\hline
	BEDXL &32  &47    &{\bf 15}  &30 &19 & 11 \\% \hline
 GRU-D   &28 &40   &12 &37  &24  & {\bf 13}  \\ %\hline
	%BM &34  &49   &{\bf 15} &28 &17 &11   \\ %\hline
	WIAED &28 &44   &16 &35  &17  &{\bf 18}  \\ \hline
\end{tabular}
%\end{center}
\end{table}

\section{Conclusions and Future Work}
\label{sec:Conclusions}
We proposed a novel lossless, compressed encoding scheme of the input window sequence using RNNs. This is equally applicable for classification and forecasting. Based on this novel scheme, we proposed an encoder-decoder framework for a general multi-step forecasting   with possible exogenous inputs. We showed that multiple interesting variations are possible on the associated decoder. We  demonstrated 
the utility of this ED framework on multiple data-sets where our proposed approach was outperforming the current state-of-art RNN baselines. In particular, our approaches demonstrated superior or comparable performances in both scenarios where (i)data was synthetically masked with varied masking window length choices (data-sets D1 and D3) and (ii)when data was naturally missing (as in D2). 
{\em Further, our experiments also indicate that our DEGD variant was the more robust variant in performance (compared to DEMI which performed poorly on D2).} We also demonstrated the applicability of our proposed architectures on both single (D1) and multi sequence data sets (D2 and D3).
As future work, we would like to validate our approach on classification tasks. We would also like to explore approaches where target imputation can be avoided (before 
model building) in forecasting applications.      

\begin{comment}
The forecasting setting allowed us to consider enhanced encoder-decoder architectures for time series prediction with 
multi-step (target) learning feature and prediction ability.
The first architecture generalized a linear Seasonal ARX model using multiple encoders, each of which captured lag correlations from one or more cycles
behind the prediction instant. The second architecture encodes  the missingness pattern in the data using two encoders  in a lossless
compressed fashion. We also proposed a greedy recursive algorithm to build background predictive models (one or at most a few)  for the multiple time
series problem when the sequence lengths are small OR when the exogneous input shows very minimal fluctuation in its values. We demonstrate extensively 
the effectiveness of both the proposed architectures on two real data sets each, where our proposed architectures did better than all the strong baselines
while it outperformed a couple of them.  As a future work, we would intend to investigate how attention mechanisms can be incorporated to these frameworks to
further enhance prediction. We  reckon that the proposed architectures can be potentially useful  in many more real world scenarios. 
\end{comment}

%\vspace{-0.1in}

%%%%%%%%%%%%%%%%%%%%%%%%%%%%%%%%%%%%%%%%%%%%%%%%%%%%

%\input{appendix}

\bibliographystyle{ACM-Reference-Format}
\bibliography{TransportationRecent}


\end{document}